\documentclass[runningheads]{llncs}
\usepackage[T1,T2A]{fontenc} 
\usepackage{graphicx}
\usepackage{color}
\usepackage[russian,english]{babel} 

\usepackage{multirow}
\usepackage{hyperref}
\usepackage{xcolor}
\begin{document}
\title{CEFR-oriented Russian Text Simplification Using Reinforcement Learning}
\titlerunning{CEFR-oriented Russian Text Simplification Using RL}
\author{Kristina Mustakova\inst{1}\orcidID{0009-0007-1803-1234} \and 
Alexey Sorokin\inst{1,2}\orcidID{0000-0003-3877-4223}}
\authorrunning{K. Mustakova, A. Sorokin}
\institute{MSU Center of Artificial Intelligence\and 
Yandex\\
\email{mustakovakn@my.msu.ru, a.sorokin@iai.msu.ru}}
\maketitle
\setcounter{footnote}{0}
\begin{abstract}
We present a study on controlled automatic simplification of Russian texts using preference-based optimization algorithms with a focus on adapting texts to the A1, A2 and B1 proficiency levels. We fine-tune YandexGPT using Rejection Sampling and Direct Preference Optimization (DPO) methods to improve control over the complexity of the generated text. We compare several training configurations that differ in dataset composition, text type and the strategy used to construct preference pairs for DPO. For automatic evaluation we train a regression model based on ruRoBERTa-large to predict continuous complexity scores corresponding to CEFR levels. We also consider semantic similarity, lexical coverage and quantitative readability metrics. Human evaluation assesses simplicity, fluency and adequacy. Our best model achieves 87.15\% target level accuracy, an MAE of 0.35 CEFR levels and a target interval MAE of 0.11. Manual assessment shows that 75\% of the evaluated simplifications match the target morphological and syntactic level, and fluency and adequacy scores are especially high for A2 and B1. Although A1 remains the most challenging level, the model significantly reduces the proportion of vocabulary and grammatical constructions beyond the target level while generally preserving meaning. Our model is freely available on Huggingface\footnote{\url{https://huggingface.co/kmustakova/RuSimplifier_DPO}}.
\end{abstract}

\section{Introduction}\label{introduction}
Text simplification is the process of reducing the linguistic complexity of a text in order to make it easier to understand and read while preserving its original meaning \cite{alva-manchego2020}. One important motivation for developing text simplification systems is the need to make texts more accessible to different groups of readers, including children, foreign language learners and people with cognitive or learning impairments. In this study, we focus specifically on Russian as a Foreign Language (RFL) learners  as the target audience for simplification.

Most research on automatic text simplification has focused on outputs that are only generally simpler than source texts, usually by reducing lexical, syntactic complexity or overall readability score. However, such approaches do not necessarily take into account the needs of specific readers. This is particularly important in educational context where texts should match the lexical, syntactic and morphological abilities expected at different proficiency levels. Controlled simplification addresses this need by producing outputs aligned with the requirements of each proficiency level rather than reducing overall complexity.

In this paper, we study CEFR-oriented automatic text simplification of Russian texts for A1, A2, B1  proficiency levels. We fine-tune YandexGPT model using Rejection Sampling and Direct Preference Optimization \cite{rafailov2023direct} and compare several preference datasets, analyzing the effect of training data size and the strategy of its construction across texts from different domains. The resulting simplifications are evaluated both automatically and manually along several criteria. The best model, trained with DPO, achieves a mean absolute error of 0.35 CEFR levels, a mean distance from the target CEFR interval of 0.11 and 87.15\% target level accuracy while effectively adapting the linguistic features of the text to the target level and preserving the original content.

\section{Related Work}
The problem of controlled text simplification remains less studied than that of general simplification. The main obstacle is the limited availability of annotated data since most existing corpora do not provide target level information. 
\subsection{Supervised Controlled Simplification}
Early approaches to automatic text simplification often treated the task as monolingual machine translation \cite{wubben2012sentence,wang2016text}. Subsequent work extended this approach by adding special tokens that specify the required level of complexity. Such control tokens have been successfully used for school grade level simplification in English \cite{scarton2018} using seq2seq NMT models. Similarly, for German, CEFR-oriented simplification was considered using Transformer-based \cite{vaswani2017attention} NMT models with source-side labels for A1, A2 and B1 levels \cite{spring2021}. 

For Russian this direction was explored by adapting the control token framework of \cite{martin2020controllable}. In particular, the approach used mBART and T5 models using tokens for compression (NbChars), paraphrasing degree (LevSim), syntactic complexity (DepTreeDepth) and CEFR level \cite{dmitrieva2023}. The results showed that NbChars and LevSim tokens worked well for Russian, while DepTreeDepth and CEFRgrade were more challenging, especially when combined with other control tokens.

\subsection{Reinforcement Learning for Controlled Simplification}
Another approach to controllable text simplification is based on reinforcement learning (RL). 

A previous study applies REINFORCE \cite{williams1992} to a Transformer simplification model with grade level tokens and optimizes it with a sentence-level reward that penalizes the distance between the target complexity and the complexity predicted by the estimation model \cite{yanamoto2022}. 

More recently, CEFR-oriented simplification for English as a Second Language (ESL) learners has been addressed by fine-tuning an instruction-tuned LLM using Proximal Policy Optimization (PPO) \cite{schulman2017proximal} without relying on a parallel data \cite{li2025aligning}. The method combines lexical and sentence-level rewards and improves lexical coverage, better aligning generated texts with the target CEFR level. 

\section{Method}
Our goal was to study automatic simplification of Russian texts for learners at A1, A2 and B1 proficiency levels. Given a complex text at C1-C2 levels and a specified target level as input, the model is expected to generate a simplified variant that matches the target level across all criteria while preserving the core meaning of the original text. The simplification process is implemented at the level of short text fragments (1.55 sentences and 24.14 words per text on average).
\subsection{Complexity Assessment Model}
Since Textometr\footnote{\url{https://textometr.ru/}} \cite{laposhina2021textometr}, an existing feature-based CEFR estimator for Russian, does not provide an API for automatic processing, we trained our own text complexity regression model\footnote{\url{https://huggingface.co/kmustakova/ru-cefr-assessor}} using Textometr's predictions as target scores.

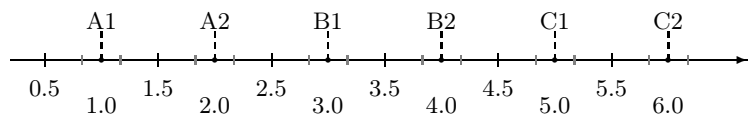
\begin{figure}[h]
    \centering
    \setlength{\unitlength}{1.5cm}
    \begin{picture}(6.5,1.0)
        \put(0.2,0.5){\vector(1,0){6.5}}
        \multiput(0.83, 0.46)(1.0,0){6}{\color{gray}\line(0,1){0.08}}
        \multiput(1.17, 0.46)(1.0,0){6}{\color{gray}\line(0,1){0.08}}
        \multiput(0.5,0.44)(1.0,0){6}{\line(0,1){0.12}}
        \multiput(1.0,0.5)(1.0,0){6}{\circle*{0.05}}
        \multiput(1.0,0.5)(0.0,0.1){3}{\line(0,1){0.05}}
        \multiput(2.0,0.5)(0.0,0.1){3}{\line(0,1){0.05}}
        \multiput(3.0,0.5)(0.0,0.1){3}{\line(0,1){0.05}}
        \multiput(4.0,0.5)(0.0,0.1){3}{\line(0,1){0.05}}
        \multiput(5.0,0.5)(0.0,0.1){3}{\line(0,1){0.05}}
        \multiput(6.0,0.5)(0.0,0.1){3}{\line(0,1){0.05}}
        \put(1.0,0.85){\makebox(0,0){A1}}
        \put(2.0,0.85){\makebox(0,0){A2}}
        \put(3.0,0.85){\makebox(0,0){B1}}
        \put(4.0,0.85){\makebox(0,0){B2}}
        \put(5.0,0.85){\makebox(0,0){C1}}
        \put(6.0,0.85){\makebox(0,0){C2}}
        \put(0.5,0.25){\makebox(0,0){0.5}}
        \put(1.0,0.1){\makebox(0,0){1.0}}
        \put(1.5,0.25){\makebox(0,0){1.5}}
        \put(2.0,0.1){\makebox(0,0){2.0}}
        \put(2.5,0.25){\makebox(0,0){2.5}}
        \put(3.0,0.1){\makebox(0,0){3.0}}
        \put(3.5,0.25){\makebox(0,0){3.5}}
        \put(4.0,0.1){\makebox(0,0){4.0}}
        \put(4.5,0.25){\makebox(0,0){4.5}}
        \put(5.0,0.1){\makebox(0,0){5.0}}
        \put(5.5,0.25){\makebox(0,0){5.5}}
        \put(6.0,0.1){\makebox(0,0){6.0}}
    \end{picture}
    \caption{Continuous CEFR scale.}
    \label{fig:scale}
\end{figure}

We consider language proficiency as a continuum. RuRoBERTa-large\footnote{\url{https://huggingface.co/ai-forever/ruRoberta-large}} \cite{zmitrovich2023family} was fine-tuned as a regression model to predict a continuous text complexity score mapped to the CEFR scale, ranging from A1 to C2. The target values were represented as real numbers from 0.5 to 6.0 (see Figure \ref{fig:scale}). Moreover, since previous studies \cite{laposhina2023elo} have highlighted the practical use of a more detailed distinction within CEFR levels, we further divide each CEFR interval into three sublevels -- beginning, middle and end -- as shown on Figure \ref{fig:scale}.

\noindent We used RuFoLa \cite{laposhina2020corpus} dataset combined with an additional dataset based on Elo ratings \cite{laposhina2023elo} for model training. After merging and filtering, the final dataset consisted of 1101 texts. 

The model was trained using AdamW optimizer for 5 epochs with a learning rate of 1e-5. The best checkpoint was selected based on the lowest MSE on the validation set. The model achieved MAE of 0.5 on the test set indicating that its predictions deviated from the reference scores by half a level on average (see Table \ref{tab:evaluator_metrics}).

\begin{table}[ht]
    \centering
    \setlength{\tabcolsep}{8pt}
    \caption{Text complexity regression metrics on the validation (at the best checkpoint) and test sets.}
    \label{tab:evaluator_metrics}
    \begin{tabular}{lccccc}
    \hline
        & MSE & RMSE & MAE & $R^2$ & Explained variance \\
        \hline
        validation & 0.39 & 0.62 & 0.45 & 0.84 & 0.87 \\
        test & 0.51 & 0.71 & 0.50 & 0.73 & 0.83 \\
        \hline
    \end{tabular}
\end{table}

\noindent The described complexity scorer was used for two purposes. First, to select candidates for constructing the preference dataset, and second, to analyze the generated simplifications on the test set. 
\subsection{Data}
\subsubsection{Source Texts}
We compiled datasets of original complex texts in Russian which served as input for the generation of simplified versions. The texts were collected from two groups of datasets. The texts of these two groups  differ both stylistically and in size.
\paragraph{Educational Learner-oriented Source Texts} The first group includes texts from the RuAdapt\footnote{\url{https://github.com/Digital-Pushkin-Lab/RuAdapt}} corpus \cite{dmitrieva2021creating} and RU-CEFR-Short\footnote{\url{https://github.com/Digital-Pushkin-Lab/RU-CEFR-Short}} dataset \cite{glazkova-etal-2026-cefr} which consists of educational texts of various genres selected from RFL textbooks. 

C-level texts were extracted from the Encyclopedic subcorpus of RuAdapt, aligned at the sentence level. The subcorpus consist of entries from Multimedia Linguistic and Cultural Dictionary\footnote{\url{https://ls.pushkininstitute.ru/}} and cover various aspects of Russian culture, as well as phenomena related to language, geography and other fields. Only sentences containing at least 15 words were included.

RU-CEFR-Short consists of short Russian text fragments (15-30 words) annotated with CEFR levels and extracted from RFL textbooks and online resources. Only texts with levels C1 and C2 were included in the dataset.

The texts from both sources were merged into a single dataset. Thus, 3760 original complex texts were obtained from the first group of sources, of which 10\% (376 texts) were used as the test set.

\paragraph{Science Source Texts} We use the Readability subcorpus of the Taiga\footnote{\url{https://tatianashavrina.github.io/taiga_site/downloads}} corpus \cite{shavrina2017methodology} based on <<N+1>> science publication covering science, technology, space, gadgets and transport. All texts were divided into short fragments containing at least 15 words, resulting in 34793 texts.
\subsubsection{Candidate Generation} The input consisted of a source complex text and a target simplification level (A1, A2 or B1). For each input pair, multiple candidate simplifications were generated using the following prompt template, where \{text\} denotes the source text and \{level\} the target level:

\textit{Тебе будет дан текст на русском языке. Твоя задача -- упростить этот текст. Упрощенный текст должен соответствовать уровню \{level\}. Обя\-зательно меняй грамматику, слова и длину предложения, но сохраняй исходное содержание.\\
ТЕКСТ: \{text\}}

\textit{(You will be given a text in Russian. Your task is to simplify the text. The simplified text must correspond to the \{level\} CEFR level. You must modify the grammar, vocabulary and sentence length, while preserving the original meaning.\\
TEXT: \{text\})}

Simplification candidates were generated using sampling decoding. For each input prompt, the model generated up to 7 hypotheses. To ensure output diversity, nucleus sampling was applied with $\mathit{top\!\_p}=0.9$ together with a sampling temperature of 1.5. From the generated candidates a preferred variant (chosen, $y_w$) and a less preferred variant (rejected, $y_l$) were selected. The selection of chosen and rejected options was performed automatically based on our estimation model scores. Thus, each training example was represented as a preference pair ($y_w$, $y_l$) where $y_w$ denotes the simplification hypothesis that best matches the target level, while $y_l$ denotes a less suitable alternative. We specify the definition of matching quality below. 

\subsubsection{Preference Data Construction} Since the chosen-rejected pairs can be formed in several ways, we considered multiple preference data construction strategies. The use of multiple configurations was motivated by the fact that typical undesirable deviations from the target level differ across CEFR levels. For A1, the only type of negative example is an overly complex candidate, i.e., a variant with an insufficient degree of simplification. However, for A2 and B1 a negative example may be either overly complex or overly simple with respect to the target level. Therefore, different preference dataset configurations were considered as a way to determine which strategy for selecting examples yields the best results. Four main configurations were constructed to compare the selection strategies.
\begin{itemize}
    \item \textbf{\textit{v0}}:
    Candidates were ranked by the distance between their predicted complexity score and the center of the target level on the continuous scale. The hypothesis with the smallest distance was selected as chosen, while the one with the largest distance was selected as rejected regardless of whether it was simpler or more complex than the target level. If multiple candidates had the same minimum distance and matched the target level, all of them were included as preferred. This was the case for 67 groups ($\approx 0.66\%$ of all groups), resulting in 88 additional pairs.
    \item \textbf{\textit{v1}}: For A1, the simplest candidate, i.e, the one with the lowest predicted complexity score, was selected as chosen, while the most complex candidate was selected as rejected. For B1, the opposite strategy was used: the candidate with the highest predicted complexity score was selected as chosen and the one with the lowest score as rejected. For A2, the same strategy as in \textit{v0} was used.
    \item \textbf{\textit{v2}}:
    For A2, the preferred candidate was selected as the one closest to 2.3 rather than to the level center of 2.0. This higher point corresponds approximately to the midpoint of the "end of A2" sublevel and was intended to reduce the selection of candidates near the A1-A2 boundary. The rejected candidate was chosen as the one deviating most from 2.3 toward lower complexity when possible. Thus, the configuration aimed to better separate A2 simplifications from A1. For A1 and B1, the same strategy as in \textit{v1} was used.
    \item \textbf{\textit{v3}}:
    Rejected candidates for B1 included both overly simple and overly complex options, aiming to achieve a 50/50 balance where possible. This configuration tested a more neutral strategy for B1. For A2, such balancing was not performed separately since the previous selection strategy already ensured an approximately balanced distribution of error directions. For A1 and A2, the same strategy as in \textit{v1} was used.
\end{itemize}

The configurations were developed sequentially: each subsequent configuration introduced a specific modification motivated by shortcomings observed in the outputs of the previous one.  

Thus, four datasets were constructed from the first group of sources, i.e., educational learner-oriented texts. They were named according to the corresponding strategy used: \textit{v0}, \textit{v1}, \textit{v2} and \textit{v3}. For science texts, two datasets were constructed: \textit{nplus1} and \textit{nplus1\_2}, both were based on \textit{v3} strategy. \textit{Nplus1} dataset was created without additional filtering of positive examples, which made it possible to preserve the maximum number of available pairs. In the second version, i.e., \textit{nplus1\_2}, we added an additional constraint on the regression score of $y_w$, so that a pair was included only if the score fell within the specified acceptable range. In addition, two merged datasets was created. The \textit{merged} dataset combines \textit{nplus1\_2} with the A1-A2 subsets of \textit{v1}. The \textit{merged\_2} dataset combines \textit{v3} with a subset of \textit{nplus1\_2} of the same size. Table \ref{tab:datasets} presents the final sizes of all datasets.
\begin{table}[ht]
    \centering
    \setlength{\tabcolsep}{14pt}
    \caption{Size of preference datasets.}
    \label{tab:datasets}
    \begin{tabular}{lcccc}
        \hline
        \textbf{Dataset} & \textbf{A1} & \textbf{A2} & \textbf{B1} & \textbf{Total} \\
        \hline
        \textit{v0} & 3424 & 3425 & 3391 & 10240 \\
        \textit{v1}, \textit{v2}, \textit{v3} & 3384 & 3384 & 3384 & 10152 \\
        \hline
        \textit{nplus1} & 34207 & 34207 & 34205 & 102619 \\
        \textit{nplus1\_2} & 13143 & 23366 & 29555 & 66064 \\
        \hline
        \textit{merged} & 16527 & 23366 & 29555 & 72832 \\
        \textit{merged\_2} & 6768 & 6768 & 6768 & 20304 \\
        \hline
    \end{tabular}
\end{table}

\noindent Table \ref{tab:yl_direction} shows clear differences in whether rejected responses are simpler ($\downarrow$) or more complex ($\uparrow$) than chosen ones in the generated preference datasets. Overall, the observed distributions are consistent with the expected behavior of the corresponding dataset construction strategies. 

\begin{table}[ht]
    \centering
    \setlength{\tabcolsep}{6pt}
    \caption{Direction of rejected responses compared with chosen responses across target levels and preference datasets (\%). $\uparrow$ and $\downarrow$ denote that the rejected response is more complex or simpler than the corresponding chosen response, respectively.}
    \label{tab:yl_direction}
    \begin{tabular}{l|cc|cc|cc}
    \hline
    & \multicolumn{2}{c}{A1} & \multicolumn{2}{c}{A2} & \multicolumn{2}{c}{B1} \\
    & $\uparrow$ & $\downarrow$ & $\uparrow$ & $\downarrow$ & $\uparrow$ & $\downarrow$ \\
    \hline
    \textit{v0} & 98.83 & 1.17 & 50.83 & 49.17 & 20.73 & 79.27 \\
    \textit{v1} & 100.0 & 0.00 & 51.77 & 48.23 & 0.00 & 100.0 \\
    \textit{v2} & 100.0 & 0.00 & 16.08 & 83.92 & 0.00 & 100.0 \\
    \textit{v3} & 100.0 & 0.00 & 51.77 & 48.23 & 30.38 & 69.62 \\
    \textit{nplus1} & 100.0 & 0.00 & 93.71 & 6.28 & 50.00 & 50.00 \\
    \textit{nplus1\_2} & 100.0 & 0.00 & 90.81 & 9.19 & 50.00 & 50.00 \\
    \textit{merged} & 100.0 & 0.00 & 85.87 & 14.13 & 50.00 & 50.00 \\
    \textit{merged\_2} & 100.0 & 0.00 & 71.47 & 28.53 & 40.25 & 59.75 \\
    \hline
    \end{tabular}
\end{table}
\subsection{Simplification Model}
Fine-tuning was performed on instruction-tuned language model YandexGPT-5-Lite-8B-instruct\footnote{\url{https://huggingface.co/yandex/YandexGPT-5-Lite-8B-instruct}} using LoRA \cite{hu2022lora} adapters with rank $r = 16$ and $\alpha = 32$. 
\subsubsection{Optimization Methods}
\paragraph{Rejection Sampling}
We used Rejection Sampling (RS) as the first method. The model was fine-tuned using the standard supervised fine-tuning (SFT) objective on pairs $(x, y_w)$  where $x$ denotes the input prompt and $y_w$ denotes the winning simplification. The rejected response $y_l$ is not used during training.
\paragraph{Direct Preference Optimization}
As a second and main approach, we applied DPO which directly optimizes the model on preference pairs. The training data consisted of triples $(x, y_w, y_l)$ where $x$ is the input prompt containing the original text and the target level, $y_w$ is the preferred simplification and $y_l$ is the less appropriate variant. DPO directly optimizes the policy model to increase the likelihood of preferred simplification over rejected one without explicitly training a separate reward model.

The model was first adapted using SFT on pairs $(x, y_w)$ in order to learn the task format. The DPO stage was performed with $\beta = 0.1$ and a learning rate of 5e-6. 
\subsubsection{Evaluation Methods}
The quality of simplification was assessed using a combination of automatic metrics and manual annotation. The evaluation focused on the degree of alignment with the target complexity level, as well as other important aspects of simplification quality.
\paragraph{Automatic Assessment} 
The main method for assessing the quality of simplifications was a complexity score assigned to each generated text by our regression model. This score was used as the primary criterion for automatic comparison of models. For the model that achieved the best results on the primary score, we conducted an additional analysis across several dimensions:
\begin{itemize}
    \item \textbf{Lexical complexity:} 
    The lexical minima of the Russian state testing system \cite{andryshina2015a1,andryshina2015a2,andryshina2015b1} provide lists of lexical units expected at each CEFR level from A1 to C1. We assessed lexical complexity by measuring the proportion of words tokens whose lemmas were not covered by the lexical minimum corresponding to the target level.
    \item \textbf{Semantic similarity:} 
    To measure meaning preservation, both original and simplified texts were encoded using FRIDA\footnote{\url{https://huggingface.co/ai-forever/FRIDA}} sentence embedding model in \textit{paraphrase} mode and cosine similarity was calculated between the resulting normalized vectors.
    \item \textbf{Quantitative metrics:} 
    We calculated several additional text characteristics, including changes in average sentence length (in words), average word length (in characters) and the number of punctuation marks per sentence. These characteristics are relevant to text complexity as has also been noted in studies for Russian \cite{laposhina2020corpus}. They are also the key parameters used in both readability formulas and state standards for RFL.
\end{itemize}
\paragraph{Human Annotation}
We randomly selected 60 texts from the test set with an equal number of examples for each target level. Manual evaluation was performed using 3 most commonly used dimensions of text simplification quality: simplicity, fluency and adequacy. 

Fluency and adequacy were evaluated by three native speakers with linguistic background. Fluency measured how understandable, grammatically correct and natural the generated text was. Adequacy measured whether the generated text preserved the core content of the original text without omitting important information, changing facts or adding unsupported information. Both criteria were assessed on a 1-5 scale where higher scores indicating better quality. Inter-annotator agreement, calculated using Krippendorff's alpha, was 0.71 and 0.70 for fluency and adequacy, respectively. 

Simplicity was evaluated separately by one annotator. The criterion focused on whether the generated text matched the target level in terms of morphology and syntax. The scale ranged from -2 to 2 where 0 indicated an exact match to the target level with respect to the morphological and syntactic material specified in the RFL State Educational Standards \cite{vladimirova2001a1,nakhabina2001a2,andryshinastandard1999b1,ivanova1999b2,ivanova1999c1}. Negative values were assigned when the text lacked constructions expected at the target level and instead corresponded to grammatical material of a lower level. Accordingly, positive values were assigned when the text contained morphological or syntactic constructions beyond the target level. For example, -1 indicated a deviation of one level below the target level whereas 1 indicated a deviation of one level above it.
\section{Results}
The test sample consisted of 752 complex source texts and covered both subsets: \textit{test\_ruadapt}, containing educational RFL texts, and \textit{test\_n+1}, containing science texts from the <<N+1>> publication. Each text was simplified into three target levels, resulting in 2256 complex-simplified text pairs for analysis. 

Table \ref{tab:RS_results} shows results for the models trained using RS. MAE$_i$ denotes interval MAE, i.e., the mean distance of the predicted complexity score from the target interval on the continuous CEFR scale; MAE denotes ordinal level MAE, i.e., the mean absolute difference between predicted and target CEFR levels after mapping the regression score to A1-C2. The method itself remains limited. The analysis of the predicted level distribution shows that models tend to shift generations toward the middle of the scale, primarily toward A2. Consequently, A2 appears to be the most stable level since generations are often assigned to it both when A2 is the target and when the model fails to match adjacent target levels. Larger deviations (exceeding one level) are especially noticeable for A1 on the \textit{test\_n+1} subset, indicating that the models may provide texts that are significantly more complex than required. The best results are achieved with merged dataset, suggesting that both source texts diversity and data size are important. 
\begin{table}[ht]
    \centering
    \setlength{\tabcolsep}{1pt}
    \caption{Rejection Sampling: target level accuracy.}
    \label{tab:RS_results}
    \begin{tabular}{l|cccccc|cccccc}
    \hline
    & \multicolumn{6}{c|}{test\_n+1} & \multicolumn{6}{c}{test\_ruadapt} \\
    \hline
    & MAE & MAE$_i$ & total & A1 & A2 & B1 & MAE & MAE$_i$ & total & A1 & A2 & B1 \\
    \hline
    base & 1.07 & 0.66 & 44.24 & 8.51 & 50.53 & 73.67 & 0.78 & 0.38 & 60.28 & 57.18 & 83.78 & 39.89 \\
    \hline
    \textit{v0} & 1.02 & 0.60 & 44.77 & 16.22 & 47.07 & 71.01 & 0.69 & 0.32 & 65.34 & 62.77 & 85.90 & 47.34 \\
    \textit{v1} & 1.07 & 0.65 & 41.31 & 14.36 & 46.01 & 63.56 & 0.71 & 0.32 & 66.58 & 64.89 & 83.78 & 51.06 \\
    \textit{v2} & 1.07 & 0.66 & 40.34 & 15.16 & 39.10 & 66.76 & 0.71 & 0.32 & 65.69 & 61.97 & 82.71 & \textbf{52.39} \\
    \textit{v3} & 1.01 & 0.60 & 44.68 & 14.10 & 50.27 & 69.68 & 0.68 & 0.32 & 66.22 & 62.23 & 84.84 & 51.60 \\
    \hline
    \textit{nplus1} & 0.87 & 0.47 & 54.52 & 20.48 & 67.82 & 75.27 & 0.65 & 0.29 & 68.44 & 71.54 & 86.44 & 47.34 \\
    \hline
    \textit{merged} & \textbf{0.73} & \textbf{0.36} & \textbf{61.35} & \textbf{32.71} & \textbf{72.61} & \textbf{78.72} & \textbf{0.63} & \textbf{0.27} & \textbf{68.97} & \textbf{72.61} & \textbf{90.43} & 43.88 \\
    \hline
    \end{tabular}
\end{table}

\noindent The results obtained using DPO are significantly higher than those obtained using RS (see Table \ref{tab:DPO_results}).

\begin{table}[!ht]
    \centering
    \setlength{\tabcolsep}{0.7pt}
    \caption{Direct Preference Optimization: target level accuracy.}
    \label{tab:DPO_results}
    \begin{tabular}{l|cccccc|cccccc}
    \hline
    & \multicolumn{6}{c|}{test\_n+1} & \multicolumn{6}{c}{test\_ruadapt} \\
    \hline
    & MAE & MAE$_i$ & total & A1 & A2 & B1 & MAE & MAE$_i$ & total & A1 & A2 & B1 \\
    \hline
    base & 1.07 & 0.66 & 44.24 & 8.51 & 50.53 & 73.67 & 0.78 & 0.38 & 60.28 & 57.18 & 83.78 & 39.89 \\
    \hline
    \textit{v0} & 1.08 & 0.66 & 43.17 & 8.78 & 46.81 & 73.93 & 0.77 & 0.38 & 59.75 & 56.65 & 83.51 & 39.10 \\
    \textit{v1} & 0.93 & 0.53 & 51.60 & 73.14 & 71.81 & 9.84 & 0.53 & 0.22 & 76.33 & 93.35 & 85.37 & 50.27 \\
    \textit{v2} & 1.14 & 0.69 & 37.23 & 77.39 & 22.07 & 12.23 & 0.55 & 0.25 & 73.14 & 93.88 & 73.40 & 52.13 \\
    \textit{v3} & 0.67 & 0.30 & 66.40 & 73.40 & 66.76 & 59.04 & 0.49 & 0.17 & 79.26 & 93.62 & 86.97 & 57.18 \\
    \hline
    \textit{nplus1} & 0.33 & 0.09 & 93.09 & 93.35 & 94.41 & \textbf{91.49} & 0.46 & 0.16 & \textbf{82.45} & 97.07 & 91.76 & \textbf{58.51} \\
    \textit{nplus1\_2} & 0.31 & \textbf{0.07} & \textbf{93.35} & 97.87 & \textbf{97.87} & 84.31 & 0.48 & 0.17 & 79.96 & 98.94 & 92.02 & 48.94 \\
    \hline
    \textit{merged} & 0.48 & 0.21 & 84.93 & 74.20 & 93.88 & 86.70 & 0.48 & 0.20 & 78.19 & 90.96 & 91.76 & 51.86 \\
    \textit{merged\_2} & \textbf{0.28} & \textbf{0.07} & 92.38 & \textbf{100.0} & 96.01 & 81.12 & \textbf{0.42} & \textbf{0.15} & 81.91 & \textbf{100.0} & \textbf{93.35} & 52.39 \\
    \hline
    \end{tabular}
\end{table}
The smaller configurations (\textit{v0-v3}) indicate that DPO strongly depends on how the preference pairs are constructed. Unbalanced strategies may lead to systematic shifts toward either oversimplification or overcomplication. In particular, \textit{v0} does not provide a sufficiently clear direction of optimization, since it ignores whether the rejected candidate falls above or below the target level and may also assign multiple preferred candidates to the same source text, resulting in poor alignment and even slightly worse simplification quality than the base model in zero-shot setting. \textit{v1} demonstrates that explicitly controlling the direction can substantially improve A1 and A2, but selecting the most complex candidate as preferred for B1 introduces a strong bias toward overcomplication and sharply reduces B1 accuracy. \textit{v2} further amplifies this effect as shifting the preferred A2 candidate toward the upper boundary of the level causes systematic overcomplication rather than better distinction between the two lowest levels. \textit{v3} uses a more balanced strategy for A2 and B1, with negative examples representing both overly simple and overly complex outputs. This yields the most consistent behavior across levels and the best overall results, suggesting that balanced preference pairs are more effective than strategies in which most negative examples are shifted in only one direction.

The \textit{nplus1} models significantly improve accuracy, particularly on the similar texts. However, their advantage is less stable when transferred to the \textit{test\_ruadapt} set. The model trained with \textit{merged\_2} dataset is particularly notable. It achieves one of the best overall results and the lowest MAE on both test sets, although the training set contains only 20304 examples, which is considerably smaller than \textit{nplus1} and \textit{nplus1\_2}. In terms of automatic assessment, B1, especially in \textit{test\_ruadapt}, shows the weakest correspondence with the target level, where error generations typically falling in the middle to upper A2 range (middle and end A2 sublevels). Therefore, \textit{merged\_2} is selected as the best model for further analysis\footnote{\url{https://huggingface.co/kmustakova/RuSimplifier_DPO}}. Examples of simplifications produced by this model are provided in the \hyperref[appendix:examples]{Appendix}.

Results on average semantic similarity indicate that, in most cases, the meaning is largely preserved, although scores vary depending on the text type (see Table~\ref{tab:cos_sim}). The data show a consistent decrease in similarity as the target level lowers, which is especially noticeable for scientific texts. The probable explanation is that, firstly, stronger simplification entails a greater semantic divergence from the original text, and secondly, that scientific texts are more difficult to simplify while preserving the original meaning. 
\begin{table}[!ht]
    \centering
    \setlength{\tabcolsep}{10pt}
    \caption{Semantic similarity across levels for the two test subsets.}
    \label{tab:cos_sim}
    \begin{tabular}{lccc}
        \hline
        \textbf{Subset} & \textbf{A1} & \textbf{A2} & \textbf{B1} \\
        \hline
        test\_ruadapt & 0.82 & 0.90 & 0.93 \\
        test\_n+1 & 0.70 & 0.83 & 0.90 \\
        \hline
    \end{tabular}
\end{table}

\noindent Since the results for the lexical and quantitative criteria differ only slightly by text type, we present them on the full test set.

\begin{figure}[!ht]
    \centering
    \includegraphics[width=0.6\linewidth]{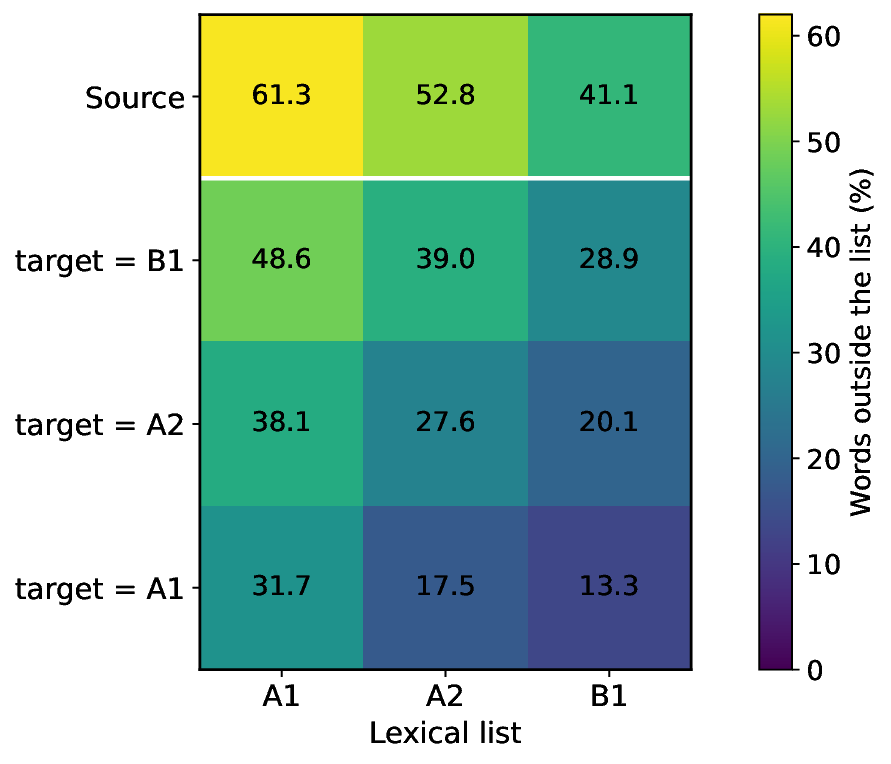}
    \caption{Percentage of words outside RFL lexical minima.}
    \label{fig:lexis}
\end{figure}

The model substantially reduces the proportion of words outside the lexical lists (see Figure \ref{fig:lexis}). The proportion decreases by 12.2\% for B1, 25.2\% for A2 and 29.6\% for A1 compared with the source texts. Thus, the lexical complexity is effectively adjusted according to the target level. It should be noted that the lexical lists do not include proper nouns or scientific terms, which are frequent in the two subsets respectively. Since all these words are often necessary for preserving the original content, they are expected to remain after simplification, which explains the nonzero proportion of words outside the lists.

\noindent The quantitative results show that simplification comes with significant text compression (see Table \ref{tab:quantitative}). All metrics gradually decrease, which suggest that the model produces structurally simpler outputs as the target level becomes lower.

\begin{figure}[!ht]
    \centering
    \includegraphics[width=0.7\linewidth]{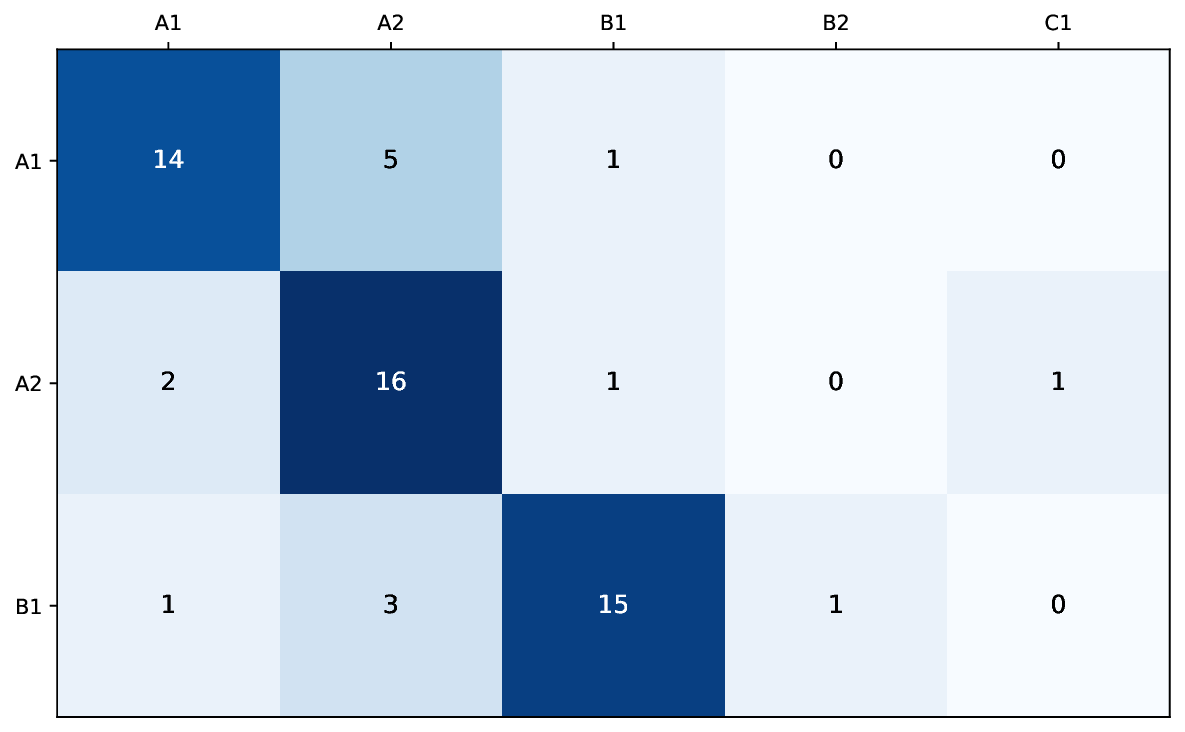}
    \caption{Confusion matrix for the manual simplicity evaluation.}
    \label{fig:simplicity_cm}
\end{figure}

\begin{table}[!ht]
    \centering
    \setlength{\tabcolsep}{6pt}
    \caption{Quantitative metrics across levels.}
    \label{tab:quantitative}
    \begin{tabular}{llrrr}
    \hline
    \textbf{Metric} & \textbf{Source} & \textbf{A1} & \textbf{A2} & \textbf{B1} \\
    \hline
    Average sentence length, words & 17.8 & 3.5 & 6.0 & 9.4 \\
    Average word length, chars & 6.37 & 4.68 & 4.97 & 5.55 \\
    Average punctuation per sentence & 3.87 & 1.15 & 1.56 & 2.15 \\
    \hline
    \end{tabular}
\end{table}
\noindent The manual simplicity evaluation reveals that the simplifications generally match the target levels with respect to morphology and syntax. Mean scores were close to 0 in most cases (see Table \ref{tab:simplicity_mean}). The best alignment was observed for A2. However, A1 simplifications for popular science subset marginally exceed the required level, suggesting that such texts may preserve morpho-syntactic constructions that are difficult to reduce fully to A1. By contrast, B1 texts in \textit{test\_ruadapt} tend to be slightly simpler. Overall, 45 out of 60 texts were assigned to the target level, giving 75\% exact matches (see Figure \ref{fig:simplicity_cm}).
\begin{table}[!ht]
    \centering
    \setlength{\tabcolsep}{10pt}
    \caption{Mean simplicity scores by target level and test subset. Scores closer to 0 indicate better alignment with the expected syntactic and morphological material, while negative and positive deviations indicate complexity below and above the target level, respectively.}
    \label{tab:simplicity_mean}
    \begin{tabular}{lccc}
         \hline
         & \textbf{A1} & \textbf{A2} & \textbf{B1} \\
         \hline
         test\_ruadapt & 0.1 & 0.0 & -0.3 \\
         test\_n+1 & 0.6 & 0.1 & -0.1\\
         \hline
    \end{tabular}
\end{table}

\noindent Test simplifications generally received high scores for fluency and adequacy, especially at A2 and B1, where mean scores were mostly above 4.7 (see Table \ref{tab:fluency_adequacy}). The lowest scores were obtained for A1 in both test sets. At this level, the model appears to simplify at the expense of fluency and adequacy, since texts become less natural, grammatical and the content is often significantly reduced.
\begin{table}[!ht]
    \centering
    \setlength{\tabcolsep}{10pt}
    \caption{Human evaluation of fluency and adequacy across target levels. Scores are presented as mean $\pm$ standard deviation, computed after averaging annotator scores for each text.}
    \label{tab:fluency_adequacy}
    \begin{tabular}{llccc}
        \hline  
         & & \textbf{A1} & \textbf{A2} & \textbf{B1} \\
         \hline
         \multirow{2}{*}{\textbf{Fluency}} 
         & test\_ruadapt & $3.08 \pm 1.04$ & $4.88 \pm 0.25$ & $4.87 \pm 0.32$ \\
         & test\_n+1 & $3.30 \pm 1.01$ & $4.27 \pm 0.60$ & $4.92 \pm 0.18$ \\
         \hline
         \multirow{2}{*}{\textbf{Adequacy}} 
         & test\_ruadapt & $3.53 \pm 0.91$ & $4.87 \pm 0.22$ & $4.87 \pm 0.22$ \\
         & test\_n+1 & $3.53 \pm 1.01$ & $4.75 \pm 0.42$ & $4.93 \pm 0.21$ \\
         \hline
    \end{tabular}
\end{table}

\section{Conclusions}
In this paper, we explored CEFR-oriented automatic simplification of Russian texts for RFL learners at A1, A2 and B1 proficiency levels. We fine-tuned YandexGPT using Rejection Sampling and Direct Preference Optimization and analyzed how the size, text domain and construction strategy of the preference dataset affect simplification quality. The best results were obtained by the model trained using DPO on the merged preference dataset. It achieved a mean absolute error of 0.35 CEFR levels, an interval MAE of 0.11 and 87.15\% target level accuracy on the full test set, according to our automatic complexity assessment model. More detailed results showed an MAE of 0.28, an MAE$_i$ of 0.07 and 92.38\% accuracy on the \textit{test\_n+1} subset, and an MAE of 0.42, an MAE$_i$ of 0.15 and 81.91\% accuracy on the \textit{test\_ruadapt} subset. 

The result suggests that the model successfully adapts texts to the required level in most cases. Automatic metrics show a consistent reduction of lexical and structural complexity as the target level decreases. Human annotation confirms that the generated texts usually correspond to the target morphological and syntactic material. Fluency and adequacy are especially high for A2 and B1 simplifications. Therefore, the two types of assessment are largely aligned. However, there is still room for improvement. A1 simplifications remain the most problematic, as the model more often loses content or produces less grammatical and natural texts. Science texts also show lower semantic similarity. These results demonstrate potential of pairwise preference optimization methods for Russian text simplification, but also emphasizes the need for further work on improving fluency and meaning preservation at the lowest proficiency levels. 

\section{Limitations}

The current evaluation has two limitations that offer directions for future work. First, although the main results of our work were confirmed by human evaluation, using the same scorer for model training and automatic quality estimation creates a risk of overfitting. As a result, improvements could partly reflect adaptation to the evaluator rather than better alignment with the target level. Second, due to resource constraints, human judgments of simplicity were provided by a single annotator, which may potentially introduce subjective bias.  

\begin{credits}
\subsubsection{\ackname}

We are grateful to Antonina Laposhina for providing the datasets used in this study and for the valuable discussions and feedback. We also thank the annotators for their contributions to the evaluation process. 

This article was prepared with the support of the Yandex.Cloud Center for Social Technologies and implemented on the Yandex Cloud platform.
\subsubsection{\discintname}
The authors have no competing interests to declare that are relevant to the content of this article.     
\end{credits}

\section*{Appendix}\label{appendix:examples}
Table \ref{tab:examples} presents one example for each source text type, simplified to three target levels. 
\begin{table}[!ht]
    \centering
    \caption{Randomly selected examples of simplifications generated by the model trained with \textit{merged\_2} dataset.}
    \label{tab:examples}
    \begin{tabular}{|p{1.15cm}|p{10.8cm}|}
         \hline
         \multicolumn{2}{|c|}{\textbf{test\_ruadapt examples}} \\
         \hline
         Original & Из-за дороговизны городской земли её застройка доходными домами была очень плотной и окна квартир, особенно дешёвых, часто выходили в тёмные дворы без деревьев, окружённые серыми стенами домов.
         
         \textit{(Due to the high cost of urban land, apartment houses were built very densely, and the windows of the flats, especially the cheaper ones, often faced dark, treeless courtyards surrounded by the grey walls of houses.)}\\
         \hline
         B1 & Из-за высоких цен на землю в городе дома строили близко друг к другу. Окна квартир, особенно недорогих, часто выходили на тёмные дворы, где нет деревьев и одни серые стены.
         
         \textit{(Due to high land prices in the city, houses were built close to each other. The windows of the flats, especially the cheaper ones, often faced dark courtyards with no trees and only grey walls.)}\\
         \hline
         A2 & В городе земля дорогая. Поэтому дома стоят близко друг к другу. Окна дешёвых квартир смотрят в тёмные дворы. Там нет деревьев, только серые стены.
         
         \textit{(Land is expensive in the city. That's why the houses are close to each other. The windows of cheap flats look out onto dark courtyards. There are no trees there, only grey walls.)}\\
         \hline
         A1 & Город дорогой. Дома стоят близко. Окна смотрят в двор. Там нет деревьев. Там серые дома.
         
         \textit{(The city is expensive. The houses are close together. The windows look out onto the courtyard. There are no trees there. There are grey houses there.)}\\
         \hline
         \multicolumn{2}{|c|}{\textbf{test\_n+1 examples}} \\
         \hline
         Original & Всего исследователи сделали 500 фотографий. Главной проблемой при изучении снимков было их низкое качество и, как следствие, небольшое количество информации, которое могло было быть получено. 
         
         \textit{(In total, the researchers took 500 photographs. The main problem in studying the images was their low quality and, as a result, the limited amount of information that could be obtained.)} \\
         \hline
         B1 & Всего учёные сделали 500 снимков. Но качество фотографий было плохое, поэтому информации с них мало получилось. 
         
         \textit{(In total, the scientists took 500 photographs. However, the quality of the photographs was poor, so a small amount of information could be obtained from them.)}\\
         \hline
         A2 & Учёные сделали 500 фото. Но снимки были плохие, поэтому информации было мало.
         
         \textit{(The scientists took 500 photos. But the images were poor, so there was little information.)}\\
         \hline
         A1 & Учёные сделали много фото. Фото плохие. На фото мало информации.
         
         \textit{(The scientists took many photos. The photos are bad. There is little information in the photos.)}\\
         \hline
    \end{tabular}
\end{table}
\bibliographystyle{splncs04}
\bibliography{refs}

@inproceedings{wubben2012sentence,
  title={Sentence simplification by monolingual machine translation},
  author={Wubben, Sander and Van Den Bosch, Antal and Krahmer, Emiel},
  booktitle={Proceedings of the 50th Annual Meeting of the Association for Computational Linguistics (Volume 1: Long Papers)},
  pages={1015--1024},
  year={2012}
}

@inproceedings{wang2016text,
  title={Text simplification using neural machine translation},
  author={Wang, Tong and Chen, Ping and Rochford, John and Qiang, Jipeng},
  booktitle={Proceedings of the AAAI Conference on Artificial Intelligence},
  volume={30},
  number={1},
  year={2016}
}

@article{alva-manchego2020,
  title={Data-driven sentence simplification: Survey and benchmark},
  author={Alva-Manchego, Fernando and Scarton, Carolina and Specia, Lucia},
  journal={Computational Linguistics},
  volume={46},
  number={1},
  pages={135--187},
  year={2020}
}

@inproceedings{spring2021,
  title={Exploring {G}erman multi-level text simplification},
  author={Spring, Nicolas and Gonzales, Annette Rios and Ebling, Sarah},
  booktitle={Proceedings of the International Conference on Recent Advances in Natural Language Processing (RANLP 2021)},
  pages={1339--1349},
  year={2021}
}

@article{vaswani2017attention,
  title={Attention is all you need},
  author={Vaswani, Ashish and Shazeer, Noam and Parmar, Niki and Uszkoreit, Jakob and Jones, Llion and Gomez, Aidan N and Kaiser, {\L}ukasz and Polosukhin, Illia},
  journal={Advances in neural information processing systems},
  volume={30},
  year={2017}
}

@inproceedings{scarton2018,
  title={Learning simplifications for specific target audiences},
  author={Scarton, Carolina and Specia, Lucia},
  booktitle={Proceedings of the 56th Annual Meeting of the Association for Computational Linguistics (Volume 2: Short Papers)},
  pages={712--718},
  year={2018}
}

@inproceedings{dmitrieva2023,
  title={Automatic text simplification of {R}ussian texts using control tokens},
  author={Dmitrieva, Anna},
  booktitle={Proceedings of the 9th Workshop on Slavic Natural Language Processing 2023 (SlavicNLP 2023)},
  pages={70--77},
  year={2023}
}

@inproceedings{martin2020controllable,
  title={Controllable sentence simplification},
  author={Martin, Louis and De La Clergerie, {\'E}ric Villemonte and Sagot, Beno{\^\i}t and Bordes, Antoine},
  booktitle={Proceedings of the twelfth language resources and evaluation conference},
  pages={4689--4698},
  year={2020}
}

@article{williams1992,
  title={Simple statistical gradient-following algorithms for connectionist reinforcement learning},
  author={Williams, Ronald J},
  journal={Machine learning},
  volume={8},
  number={3},
  pages={229--256},
  year={1992},
  publisher={Springer}
}

@inproceedings{yanamoto2022,
  title={Controllable text simplification with deep reinforcement learning},
  author={Yanamoto, Daiki and Ikawa, Tomoki and Kajiwara, Tomoyuki and Ninomiya, Takashi and Uchida, Satoru and Arase, Yuki},
  booktitle={Proceedings of the 2nd conference of the asia-pacific chapter of the association for computational linguistics and the 12th international joint conference on natural language processing (volume 2: Short papers)},
  pages={398--404},
  year={2022}
}

@inproceedings{li2025aligning,
  title={Aligning sentence simplification with {ESL} learner’s proficiency for language acquisition},
  author={Li, Guanlin and Arase, Yuki and Crespi, Noel},
  booktitle={Proceedings of the 2025 Conference of the Nations of the Americas Chapter of the Association for Computational Linguistics: Human Language Technologies (Volume 1: Long Papers)},
  pages={492--507},
  year={2025}
}

@article{schulman2017proximal,
  title={Proximal policy optimization algorithms},
  author={Schulman, John and Wolski, Filip and Dhariwal, Prafulla and Radford, Alec and Klimov, Oleg},
  journal={arXiv preprint arXiv:1707.06347},
  year={2017}
}

@inproceedings{rafailov2023direct,
 author = {Rafailov, Rafael and Sharma, Archit and Mitchell, Eric and Manning, Christopher D and Ermon, Stefano and Finn, Chelsea},
 booktitle = {Advances in Neural Information Processing Systems},
 editor = {A. Oh and T. Naumann and A. Globerson and K. Saenko and M. Hardt and S. Levine},
 pages = {53728--53741},
 publisher = {Curran Associates, Inc.},
 title = {Direct {P}reference {O}ptimization: Your Language Model is Secretly a Reward Model},
 volume = {36},
 year = {2023}
}

@inproceedings{dmitrieva2021creating,
  title={Creating an aligned {R}ussian text simplification dataset from language learner data},
  author={Dmitrieva, Anna and Tiedemann, J{\"o}rg},
  booktitle={Proceedings of the 8th Workshop on Balto-Slavic Natural Language Processing},
  pages={73--79},
  year={2021}
}

@misc{shavrina2017methodology,
  title={To the methodology of corpus construction for machine learning: {\guillemotleft}{T}aiga{\guillemotright} syntax tree corpus and parser},
  author={Shavrina, Tatiana and Shapovalova, Olga},
  year={2017},
  publisher={Saint-Petersbourg}
}

@misc{zmitrovich2023family,
      title={A Family of Pretrained Transformer Language Models for {R}ussian}, 
      author={Dmitry Zmitrovich and Alexander Abramov and Andrey Kalmykov and Maria Tikhonova and Ekaterina Taktasheva and Danil Astafurov and others},
      year={2023},
      eprint={2309.10931},
      archivePrefix={arXiv},
      primaryClass={cs.CL}
}

@article{laposhina2020corpus,
  title={A corpus of {R}ussian textbook materials for foreign students as an instrument of an educational content analysis},
  author={Laposhina, Antonina},
  journal={Russian Language Abroad},
  volume={6},
  number={283},
  pages={22--28},
  year={2020}
}

@inproceedings{laposhina2023elo,
  title={Text complexity as a non-discrete value: Russian {L}2 text complexity dataset annotation based on {E}lo rating system},
  author={Laposhina, Antonina},
  booktitle={Proceedings of the International Conference “Dialogue},
  volume={2023},
  year={2023}
}

@article{hu2022lora,
  title={Lora: Low-rank adaptation of large language models.},
  author={Hu, Edward J and Shen, Yelong and Wallis, Phillip and Allen-Zhu, Zeyuan and Li, Yuanzhi and Wang, Shean and Wang, Liang and Chen, Weizhu and others},
  journal={Iclr},
  volume={1},
  number={2},
  pages={3},
  year={2022}
}

@book{andryshina2015a1,
    author = {Andryshina, N. P. and Kozlova, T. V.},
    title = {Lexical minimum of {R}ussian as a foreign language. Level A1. Common language},
    edition = {5},
    publisher = {Zlatoust},
    adress = {St. Petersburg},
    year = {2015},
    pages = {80},
    note = {(In Russian)}
}

@book{andryshina2015a2,
    author = {Andryshina, N. P. and Kozlova, T. V.},
    title = {Lexical minimum of {R}ussian as a foreign language. Level A2. Common language},
    edition = {5},
    publisher = {Zlatoust},
    adress = {St. Petersburg},
    year = {2015},
    pages = {116},
    note = {(In Russian)}
}

@book{andryshina2015b1,
    author = {Andryshina, N. P. and Bitekhtina, G. A. and Klobukova, L. P. and Noreyko, L. N. and Odintsova, I. V.},
    title = {Lexical minimum of {R}ussian as a foreign language. Level B1. Common language},
    edition = {7},
    publisher = {Zlatoust},
    adress = {St. Petersburg},
    year = {2015},
    pages = {200},
    note = {(In Russian)}
}

@book{vladimirova2001a1,
    author = {Vladimirova, T. V. and Nakhabina, M. M. and Soboleva, N. I. and Andryshina, N. P.},
    title = {State Standard for {R}ussian as a Foreign Language. Level A1},
    edition = {2},
    publisher = {Zlatoust},
    adress = {Moscow and St. Petersburg},
    year = {2001},
    pages = {28},
    note = {(In Russian)}
}

@book{nakhabina2001a2,
    author = {Nakhabina, M. M. and Soboleva, N. I. and Starodub, V. V. and Stepanenko, V.A. and Andryshina, N. P. and Antonova, V. E. and others},
    title = {State Standard for {R}ussian as a Foreign Language. Level A2},
    edition = {2},
    publisher = {Zlatoust},
    adress = {Moscow and St. Petersburg},
    year = {2001},
    pages = {32},
    note = {(In Russian)}
}

@book{andryshinastandard1999b1,
    author = {Andryshina, N. P. and Bitekhtina, G. A. Ivanova, A. S. and Klobukova, L. P. and Krasilnikova, L. V. and Nakhabina, M. M. and others},
    title = {State Educational Standard for {R}ussian as a Foreign Language. Level B1. Common Language.},
    publisher = {Zlatoust},
    adress = {Moscow and St. Petersburg},
    year = {1999},
    pages = {36},
    note = {(In Russian)}
}

@book{ivanova1999b2,
    author = {Ivanova, A. S. and Popova, T. I and Rogova, K. A. and Yurkov, E. E},
    title = {State Educational Standard for {R}ussian as a Foreign Language. Level B2. Common Language.},
    publisher = {Zlatoust},
    adress = {Moscow and St. Petersburg},
    year = {1999},
    pages = {40},
    note = {(In Russian)}
}

@book{ivanova1999c1,
    author = {Ivanova, A. S. and Popova, T. I and Rogova, K. A. and Yurkov, E. E},
    title = {State Educational Standard for {R}ussian as a Foreign Language. Level C1. Common Language.},
    publisher = {Zlatoust},
    adress = {Moscow and St. Petersburg},
    year = {1999},
    pages = {44},
    note = {(In Russian)}
}

@inproceedings{glazkova-etal-2026-cefr,
    title = "{CEFR} Level Prediction for Short {R}ussian {L}2 Texts: Evaluating Classifiers and Instruction-Based {LLM}s",
    author = "Glazkova, Anna  and
      Laposhina, Antonina  and
      Morozov, Dmitry",
    editor = "Piperidis, Stelios  and
      Bel, N{\'u}ria  and
      van den Heuvel, Henk  and
      Ide, Nancy  and
      Krek, Simon  and
      Toral, Antonio",
    booktitle = "Proceedings of the Fifteenth Language Resources and Evaluation Conference",
    month = may,
    year = "2026",
    address = "Palma de Mallorca, Spain",
    publisher = "ELRA Language Resource Association",
    url = "https://aclanthology.org/2026.lrec-1.84/",
    doi = "10.63317/27p9pbh4oods",
    pages = "1081--1091"
}

@article{laposhina2021textometr,
  title={Textometr: an online tool for automated complexity level assessment of texts in {R}ussian as a foreign language.},
  author={Laposhina, Antonina Nikolaevna and Lebedeva, Maria Yuryevna},
  journal={Russian Language Studies},
  volume={19},
  number={3},
  pages={331--345},
  year={2021},
  note={(In Russian)}
}
\end{document}